\documentclass[conference]{ieeeconf}
\usepackage{placeins}
\usepackage[pdftex]{graphicx}
\usepackage{color}
\usepackage{transparent}
\graphicspath{{./figs/}{./jpeg/}}
\DeclareGraphicsExtensions{.pdf,.jpeg,.png}

\usepackage[cmex10]{amsmath}
\usepackage{bm,upgreek}
\usepackage{amsfonts}
\usepackage{wasysym}

\usepackage{amsthm}
\usepackage{amssymb}
\usepackage{import}
\usepackage{booktabs}
\usepackage{transparent}
\theoremstyle{plain}

\newtheorem{proposition}{Proposition}

\theoremstyle{definition}

\theoremstyle{remark}

\usepackage{import}
\usepackage{algorithmic}

\usepackage{array}

\usepackage{url}
\DeclareMathOperator{\Parallel}{Parallel}
\DeclareMathOperator{\Series}{Series}
\begin{document}
\newcommand*{\htwo}{\mathcal H_2}
\newcommand*{\hinf}{\mathcal H_\infty}
\newcommand*{\jo}{(j\omega)}
\newcommand*{\s}{(s)}
\newcommand*{\tr}{\mathrm{tr}}

\title{\bf Compliance Shaping for Control of Strength Amplification Exoskeletons with Elastic Cuffs}
\author{Gray Cortright Thomas, Jeremiah M. Coholich, and Luis Sentis% <-this % stops a space
\thanks{This work was supported by NASA Space Technology Research Fellowship grant NNX15AQ33H, ``Controlling Robots with a Spring in Their Step,'' for which Gray is the Fellow and Luis is the Advising Professor. Support also comes from the DOD. We would like to thank Apptronik Systems for providing our hardware. G.C.T and J.C. are with the Department of Mechanical Engineering, and L.S. is with the Department of Aerospace Engineering,
        University of Texas at Austin, Austin, TX 78712, USA. Send correspondence to 
        {\tt\small gray.c.thomas@utexas.edu}}%
}
%\author{Michael~Shell,~\IEEEmembership{Member,~IEEE,}
%        John~Doe,~\IEEEmembership{Fellow,~OSA,}
%        and~Jane~Doe,~\IEEEmembership{Life~Fellow,~IEEE}% <-this % stops a space
%\thanks{M. Shell was with the Department
%of Electrical and Computer Engineering, Georgia Institute of Technology, Atlanta,
%IEEE Transactions on Automatic Control,~Vol.~X, No.~X, January~2017
\markboth{Pre-Print Manuscript}{Thomas \MakeLowercase{\textit{et al.}}: Shaping Series Elastic Actuator Compliance}

\maketitle

\IEEEpeerreviewmaketitle
%the aim of this work is to find a model which explains 
\begin{abstract}
Exoskeletons which amplify the strength of their operators can enable heavy-duty manipulation of unknown objects. However, this type of behavior is difficult to accomplish; it requires the exoskeleton to sense and amplify the operator's interaction forces while remaining stable. But, the goals of amplification and robust stability when connected to the operator fundamentally conflict. As a solution, we introduce a design with a spring in series with the force sensitive cuff. This allows us to design an exoskeleton compliance behavior which is nominally passive, even with high amplification ratios. In practice, time delay and discrete time filters prevent our strategy from actually achieving passivity, but the designed compliance still makes the exoskeleton more robust to spring-like human behaviors. Our exoskeleton is actuated by a series elastic actuator (SEA), which introduces another spring into the system. We show that shaping the cuff compliance for the exoskeleton can be made into approximately the same problem as shaping the spring compliance of an SEA. We therefore introduce a feedback controller and gain tuning method which takes advantage of an existing compliance shaping technique for SEAs. We call our strategy the ``double compliance shaping'' method. With large amplification ratios, this controller tends to amplify nonlinear transmission friction effects, so we additionally propose a ``transmission disturbance observer" to mitigate this drawback. Our methods are validated on a single-degree-of-freedom elbow exoskeleton.
\end{abstract}

%% Non-writing:
%	Remaking figures in software
%	Looking up references
%	Formatting
%	Drawing new figures in Inkscape
%% Feels like writing?
%	Looking up, reading the references, and taking notes
%% Definitely writing:
%	Writing new paragraphs
%% Probably writing:
%	Editing writing
%	Writing rambling comments which eventually become section outlines
%	Anything which requires assuming the author persona
%	Deciding what to call things

\section{Introduction} \label{sec:intro}
%% Introductions should be about remembering literature facts, not BSing.

%%% Exoskeltons -> Payload, Strength, Endurance -> Strength Amplification -> Similarity to force feedback human interaction -> This paper

%% Exoskeletons
%\IEEEPARstart{E}{xoskeletons}
Exoskeletons are a broad category of wearable collaborative robots that have been successful in a wide range of applications. Some aim to recover locomotion capability lost to disease \cite{KwaNoordenMisselCraigPrattNeuhaus2009ICRA,HaribHereidAgrawalGurrietFinetBoerisDuburocqMungaiMasselinAmesSreenathGrizzle2018CSM} or offload the strenuous work of rehabilitation therapy from therapists \cite{SugarHeEA2007TNSRE,KimDeshpande2017IJRR}. Others aim to minimally detract from the performance of healthy operators while augmenting their motion with simple pre-programmed boosts to improve efficiency in walking or other predictable tasks \cite{LeeKimBakerLongKaravasMenardGalianaWalshJNR2018,ZhangFiersWitteJacksonPoggenseeAtkesonCollins2017Science}. However, different control techniques are necessary to allow exoskeletons to increase the payload, strength, and endurance of operators (healthy or otherwise) as they attempt non-repetitive, unpredictable tasks. We call such systems ``amplification exoskeletons'', since their goal is the amplification of their human operators.

% Cite binghan's IROS paper
%% Payload, Strength, Endurance
We can subdivide amplification into three types: payload amplification seeks to reduce the burden of a modeled payload on the operator; strength amplification seeks to amplify the human with respect to unmodeled loads; and endurance amplification seeks to help the operator move their own body. These three areas of amplification are all independently valuable, and most exoskeleton systems in the literature have focused on one to the exclusion of others. The BLEEX exoskeleton focuses on payload amplification, using gravity compensation and positive acceleration feedback to help the operator move a carefully modeled load \cite{KazerooniRacineHuangSteger2005ICRA,Kazerooni2005IROS}. Endurance amplification has been addressed by extending inertia reduction \cite{KongTomizuka2009TMech} and gravity compensation \cite{KongMoonJeonTomizuka2010TMech} to (reduce) the inertia and gravity of operators themselves. This paper, however, focuses on the control difficulties of strength amplification (with regard to unknown loads), in which the environmental forces can still be perceived by the operator.

%% Test Exoskeleton
\begin{figure}%
	\centering%
\resizebox{\columnwidth}{!}{%
	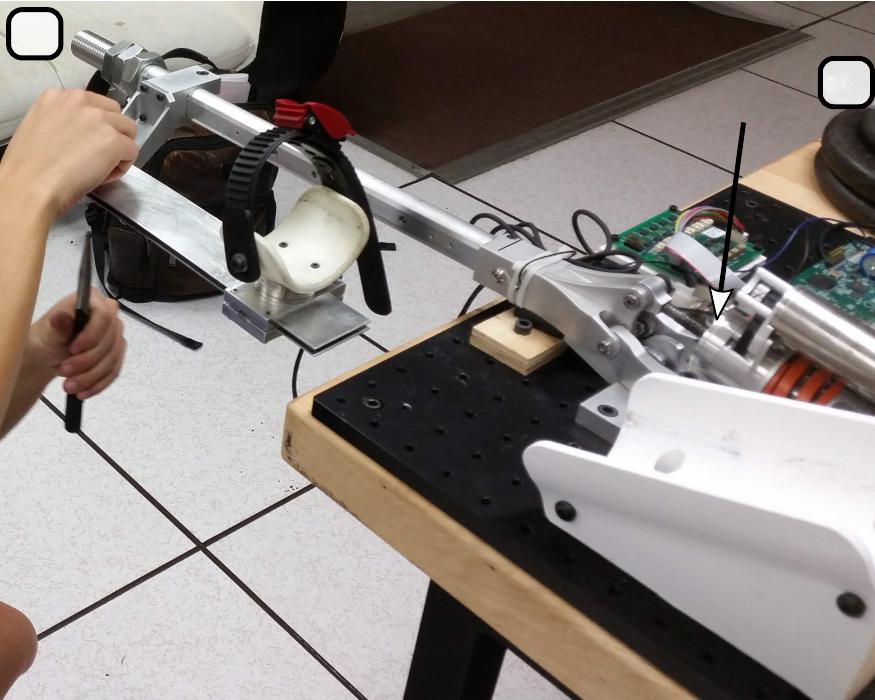}
	\caption{The experimental compliant cuff exoskeleton, assembled at the Human Centered Robotics Lab in the University of Texas at Austin, comprising: (a) a motor suspension spring and (b) a belt-drive to ball screw transmission driven by an electric motor (these two comprising an Appronik Systems Orion Series Elastic Actuator (SEA)); (c) a mount for adding weights to the exoskeleton structure, (d) a flexure spring made of two aluminum plates, (e) a six axis ATI Mini40 force/torque sensor, (f) a cuff (padding not shown) to be attached to the human forearm, and (g) an inclined arm rest to position the elbow near the kinematic rotation axis of the exoskeleton (padding typically added below the elbow for comfort).} \label{fig:hardware}
\end{figure}

% Strength Amplification
To increase the strength of an operator means to augment the interaction forces they exert on the unknown world. Amplification exoskeletons do this by being physically sandwiched between the user and the environment, measuring the forces the user applies to the exoskeleton, and applying additional forces to make the force between the exoskeleton and the environment much larger than the force between the human and the exoskeleton. As shown in the HARDIMAN I project, this is a challenging control problem \cite{MakinsonBodineFitck1969Techreport}. The system's stability depends on the human and environmental impedance \cite{Kazerooni1990TSMC}. While at first it may seem that strength and payload amplification serve the same purpose (indeed strength augmentation helps carry large, unmodeled payloads) strength amplification is a relatively impractical way to handle known payloads and is therefore a relatively unusual approach.

% Admittance control
A strategy known as admittance control, or ``getting out of the way'' \cite{LecoursStongeGosselin2012ICRA, YuRosen2013TCyb, FontanaVertechyMarcheschiSalsedoBergamasco2014RAM, JacobsenOlivier2014Patent}, uses position control on the exoskeleton and shifts a position set-point based on the human forces. This strategy is more focused on payload than strength amplification, since it requires modification and additional sensors in order to allow the operator to control the interaction force when interacting with unmoving environments \cite{KazerooniGuo1993JDSMC}. 

%While \cite{Kazerooni1990TSMC} has two sets of force sensors on the cuffs and end-effector, our approach uses the built-in force sensors of series elastic actuators and our sensors are therefore at the cuffs and actuators. This configuration (when scaled up to a whole robot) should allow whole body amplification even without knowledge of the contact location between the world and the exoskeleton.

% Similarity to force feedback human interaction
Haptic impedance controllers  \cite{ColgateHogan1988IJC,Hogan1989ICRA,ColgateBrown1994ICRA,AdamsHannaford1999TRA} share many of the control difficulties associated with strength amplification. This community has recognized the importance of the human impedance---which can change due to antagonistic muscle co-contraction \cite{Hogan1984TAC}. While many interaction controllers are designed to be energetically passive---requiring net energy from the human if it is forced to follow any cyclic trajectory---this is a conservative requirement. More accurate bounds on the human's capabilities lead to higher performance controllers which are still stable when interfacing with humans \cite{BuergerHogan2007TRO}. Bounded uncertain human impedance models have also been applied to design the closed loop interface dynamics of an amplification exoskeleton \cite{HeThomasPaineSentis2018ArXiv}.
% However this work explores the compliance shapes which can be achieved, and is not concerned with modeling any human. has so far not explored any mechanical optimizations of the cuff sensing system nor control optimization including the low level SEA controller . % add new paper

% This paper
This paper introduces a mechanical spring between the human cuff and the exoskeleton structure in order to improve the passivity properties of the system. We do not seek complementary stability with any particular bounded uncertain human model, but rather a qualitative change in the types of passive human behaviors that can destabilize the system. While the exoskeleton is fundamentally non-passive in that it amplifies the human (and adds energy in doing so), its impedance (and dynamic compliance\footnote{Dynamic compliance is a position per force transfer function, and the integral of admittance.}) at the human interface can be shaped to a considerable degree. We apply the recently developed theory of compliance shaping \cite{ThomasMehlingHolleySentisXXXX} for series elastic actuators twice to design the feedback gains for the exoskeleton. We call this controller design approach ``double compliance shaping''. We treat the exoskeleton with the elastic cuff as a series elastic actuator whose motor subsystem is itself a series elastic actuator. The contributions of the paper are in 1) the theory of double compliance shaping for exoskeleton control, 2) a transmission disturbance observer which is designed to remove nonlinear transmission effects, and 3) empirical validations of the first two contributions. Overall, this study aims to offer a framework for devising controllers and compliant cuffs for well behaved strength amplifying exoskeletons. 
\section{Modeling}\label{sec:model}
The model shown in Fig.~\ref{fig:schematics} is a series-elastic-actuator-driven exoskeleton with a series elastic cuff. While the series elastic joint (interface $\tau_j$--$\theta_j$) is a fourth order system, the actuator itself (interface $\tau_s$--$\theta_j$ ) is only second order. Motor dynamics are approximated with inertia and linear damping. No model of imperfect backdrivability is used (backdrivability will be enforced later using the disturbance observer). While the system is part prismatic and part rotational, we use a linear schematic for the whole system for simplicity. Later we employ a gain scheduling technique to keep the controller behavior similar even as the transmission ratio changes.

\subsection{Open Loop Linear Model}
%% introduce symbols glossary
\begin{table}\vspace{.1in}
	\caption{Symbol Glossary}\label{tab:notation}
	\centering\resizebox{\columnwidth}{!}{%
	\begin{tabular}{rl}
		\toprule
		Symbol & Meaning\\
		\midrule
		$\theta_m(s)$ & reflected motor deflection angle (signal)\\
		$\tau_m(s)$ & reflected motor input torque (signal)\\
		$J_m$, $B_m$ & reflected motor inertia and damping (constant)\\
		$\theta_s(s)$, $\tau_s(s)$ & spring deflection angle and torque (signal)\\
		$K_s$, $K_c$ & SEA spring and exo cuff stiffnesses (constant)\\
		$\theta_j(s)$, $\tau_j(s)$ & exoskeleton joint angle, external torque (signal)\\
		$J_j$ & exoskeleton joint inertia (constant)\\
		$\theta_c(s)$, $\tau_c(s)$ & cuff spring deflection and torque (signal)\\
		$K_1,\ B_1,\ K_2,\ B_2$ & SEA controller parameters (constant)\\
		$\tilde K_1,\ \tilde B_1,\ \tilde K_2,\ \tilde B_2$ & SEA compliance shape parameters (constant)\\
		$\hat K_1,\ \hat B_1,\ \hat K_2,\ \hat B_2$ & meta-SEA (cuff) controller parameters (constant)\\
		$\zeta,\  \hat \zeta$ & zero-pair damping ratios for SEA and meta-SEA\\
		$\hat B_m,\ \hat J_m$ & virtual motor damping and inertia (constant)\\
		$\alpha$ & amplification ratio (constant)\\
		$G_\theta(s)$& motor position feedback (transfer function)\\
		$G_s(s)$& spring torque feedback (transfer function)\\
		$G_c(s)$& cuff torque feedback (transfer function)\\
		$G_v(s)$& virtual motor input filter (transfer function)\\
		$\hat C_5(s)$ & virtual motor system compliance (trans. function)\\
		$\hat \tau_m$ & virtual motor torque (signal)\\
%		$S_i=(C_i(s),\ H_i(s))$& the i$^\mathrm{th}$ system (pair of transfer functions)\\
		\bottomrule
	\end{tabular}}
%	\rule{\columnwidth}{3pt}
\end{table}

\begin{figure}[t]%
	\centering\resizebox{\columnwidth}{!}{%
	{\footnotesize%
		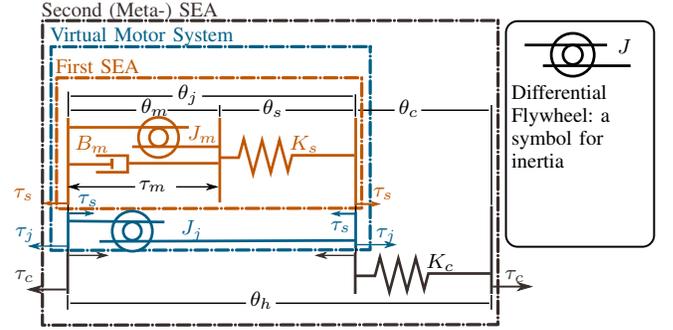%
	}}
	\caption{Double-SEA model of a compliant exoskeleton. Here the inner SEA can be tuned via compliance shaping in order to have the desired dominant second order pole locations when combined with the joint inertia. Together, these are then approximated as second order for the purpose of designing the controller of the second (meta-) SEA. A virtual motor torque input signal is adjusted by an appropriately chosen filter. Compliance shaping can therefore be applied to tuning the behavior of the meta-SEA, just like any other SEA. }\label{fig:schematics}
\end{figure}

Fig.~\ref{fig:schematics} and Tab.~\ref{tab:interconnections} describe the actuator as a sequence of series and parallel interconnections of elements with different dynamic compliances. There are seven systems here, including the feedback controllers for each of the three sensed signals: motor position $\theta_m$, spring torque $\tau_s$, and cuff torque $\tau_c$ (all reflected into the joint space using the transmission Jacobian). These seven systems are built up as follows: 1) the passive motor compliance, 2) the closed loop motor system with position feedback, or system 1's parallel interconnection with motor position feedback, 3) 2's series interconnection with spring torque feedback, 4) 3's series interconnection with the SEA spring, 5) 4's parallel interconnection with the exoskeleton, 6) 5's series interconnection with cuff torque feedback, and 7) 6's series interconnection with the cuff spring.

At each level of interconnection, we keep track of not only the system compliance, but also the effect of the motor current on the output position. We define a ``system'' to be a pair of an ``external'' compliance (position per external torque) and a ``motor'' compliance (position per motor torque), both of which are transfer functions. At each level of interconnection the meaning of position, external force, and motor torque change slightly as more components are grouped together, as shown in Tab.~\ref{tab:interconnections}.

\begin{table*}\vspace{.1in}
	\caption{Seven System Interconnections Model of a Series Elastic Exoskeleton With a Compliant Cuff}\label{tab:interconnections}\centering\resizebox{\textwidth}{!}{%
	\begin{tabular}{clccc}
		\toprule
		System (Compliance Pair) & Definition & Position & External Torque & Motor Torque\\
		\midrule
		$S_1 = (C_1(s),\ H_1(s))$ & $S_1=(\frac{1}{J_m s^2+ B_m s},\ \frac{e^{-s T}}{J_m s^2+ B_m s})$ & $\theta_m$ & $\tau_s$ & $\tau_m$\\
		$S_2 = (C_2(s),\ H_2(s))$ & $S_2 = \Parallel(S_1, [-1/G_\theta(s)C_1(s)H_1(s)])$  & $\theta_m$ & $\tau_s$ & $\tau_2 = \tau_m - G_\theta(s) \theta_m$\\
		$S_3 = (C_3(s),\ H_2(s))$ & $S_3= \Series\left(S_2 , [G_s(s)H_2(s)]\right)$ & $\theta_m$ & $\tau_s$ & $\tau_3 = \tau_2 - G_s(s) \tau_s$\\
		$S_4 = (C_4(s),\ H_2(s))$ & $S_4= \Series\left(S_3 , [1/K_s]\right)$ & $\theta_j=\theta_m+K_s^{-1}\tau_s$ & $\tau_s$ & $\tau_3$\\
		$S_5 = (C_5(s),\ H_5(s))$ & $S_5= \Parallel\left(S_4 , [1/(J_j s^2)]\right)$ & $\theta_j$ & $\tau_j=\tau_s + J_j s^2 \theta_j$ & $\tau_3$\\
		$\hat S_5 = (\hat C_5(s),\ \hat C_5(s))$ & $\hat S_5= \left(1/(\hat J_m s^2+\hat B_m s) ,\ 1/(\hat J_m s^2+\hat B_m s) \right)$ & $\theta_j$ & $\tau_j$ & $\hat \tau_m = G_v^{-1}(s) \tau_3$\\
		$S_6 = (C_6(s),\ H_5(s))$ & $S_6= \Series\left(S_5 , [G_c(s)H_5(s)]\right)$ & $\theta_j$ & $\tau_c=\tau_j$ & $\tau_6=\tau_3 - G_c(s) \tau_c$\\
		$S_7 = (C_7(s),\ H_5(s))$ & $S_7= \Series\left(S_6 , [1/K_c]\right)$ & $\theta_h=\theta_j+K_c^{-1}\tau_c$ & $\tau_c$ & $\tau_6$\\
		\bottomrule
		\end{tabular}}
\end{table*}

We then use the following two propositions to define the behavior of series and parallel interconnections.
\begin{proposition}[Systems in Parallel]
	The parallel interconnection of the system $S_1=\left(C_1(s),\ H_1(s)\right)$ and the compliance $C_2(s)$, denoted $S=\Parallel(S_1,\ C_2(s))$, is
	\begin{align}
	S=\big(&[C_1^{-1}(s) +C_2^{-1}(s) ]^{-1},\nonumber\\ &[C_1^{-1}(s) +C_2^{-1}(s)]^{-1} C_1^{-1}(s) H_1(s)\big)\end{align}
\end{proposition}

\begin{proposition}[Systems in Series]
	The series interconnection of the system $\left(C_1(s),\ H_1(s)\right)$ and the compliance $C_2(s)$, denoted $S=\Series(S_1,\ C_2(s))$, is
	\begin{align}
	S=\big([C_1(s)+C_2(s)],\  H_1(s)\big)
	\end{align}
\end{proposition}

It is also possible to represent feedback control as a virtual parallel or series system, using the following two propositions.

\begin{proposition}[Virtual Parallel Systems]
	The system $\left(C_1(s),\ H_1(s)\right)$ under the position controller $G(s)$ is equivalent to the parallel interconnection of that system and the virtual parallel system
	\begin{align}
	C^\prime(s) = -\frac{C_1(s)}{H_1(s)G(s)}
	\end{align}
\end{proposition}

\begin{proposition}[Virtual Series Systems]
	The system $\left(C_1(s),\ H_1(s)\right)$ under the force controller $G(s)$ is equivalent to the series interconnection of that system and the virtual series system
	\begin{align}
	C^\prime(s) = G(s) H_1(s)
	\end{align}
\end{proposition}

These propositions can be used to model both the compliance and output compliance of an arbitrarily complex assemblage of passive elements and sensor feedbacks. Tab.~\ref{tab:interconnections} demonstrates their application to the series elastic exoskeleton with the compliant cuff. Note that the time delay $e^{-sT}$ in the first system's input compliance is neglected in finding the nominal model. This time delay will later be critical for finding the performance limits.

The angles, torques, and dynamics in Tab.~\ref{tab:notation} (the symbol glossary) are all mechanically reflected through the applicable gear ratios into the reference frame of the joint, such that the frequency domain signals\footnote{This paper's notation neglects the explicit functional dependence on the Laplace variable~$s$ for the lower-case greek letter variables, which represent signals, except where they are introduced in Tab.~\ref{tab:notation}.} are related as follows:
\begin{gather}
\theta_j = \theta_m + \theta_s, \qquad \tau_j = J_j s^2 \theta _j + \tau_s, \nonumber\\
\tau_s = K_s \theta_s, \qquad (J_m s^2 + B_m s) \theta_m = \tau_s + \tau_m,\nonumber\\
\theta_h = \theta_j + \theta_c, \qquad \tau_j=\tau_c=K_c\theta_c.\label{eq:model}
\end{gather}
The six equations above are a complete description of this simple linear model.

\subsection{Compliance Shaping for SEAs}\label{sec:compliance_shaping}
%% introduce full state feedback controller parameterization. Wall of math
Compliance shaping for SEAs is a control strategy that dictates the gains of a controller based on the desired closed loop SEA compliance. This builds on our earlier work which introduced compliance shaping and applied it to the design of high-stiffness position control of SEAs \cite{ThomasMehlingHolleySentisXXXX}. The controller (which is also closely related to the full-state feedback controller of \cite{AlbuschafferHirzinger2000IROS}) is
\begin{equation}
	\tau_m = -(K_1 + B_1 s) \theta_m + (K_2 + B_2 s) \theta_s,
\end{equation}
with controller gain variables $K_1$, $B_1$, $K_2$, and $B_2$. 

Using the parallel and series interconnections of Tab.~\ref{tab:interconnections}, with $G_\theta(s)=-(K_1 + B_1 s)$ and $G_s(s) =  (K_2 + B_2 s)$ yields the SEA compliance 
\begin{align}
\frac{\theta_j}{\tau_s} &= \frac{J_m s^2 + (B_m+B_1+B_2) s +(K_s+K_1+K_2)}{(J_m s^2 + (B_m+B_1) s +K_1)K_s},\label{eq:long_actuator_compliance}
\end{align}
which we can rewrite in terms of a ratio of second order polynomials 
\begin{equation}
\frac{\theta_j}{\tau_s}= C_4(s) = \frac{s^2 + \tilde B_2 s +\tilde K_2}{(s^2 + \tilde B_1 s +\tilde K_1)K_s},\label{eq:motspring_compliance}
\end{equation}
in order to extract the gains of the controller as a function of the shape of the compliance:
%\begin{gather}
%\tilde K_1 = K_1/J_m, \quad \tilde K_2 = (K_s+K_1+K_2)/J_m, \nonumber\\
%\tilde B_1 = (B_m + B_1)/J_m, \quad
%\tilde B_2 =( B_m + B_1 + B_2)/J_m.\label{eq:hyper_params}
%\end{gather}
\begin{gather}
K_1=J_m \tilde K_1, \quad K_2 =J_m (\tilde K_2-\tilde K_1)-K_s, \nonumber\\
B_1=J_m \tilde B_1-B_m, \quad
B_2 =J_m (\tilde B_2-\tilde B_1).\label{eq:extracted}
\end{gather}
Thus we consider this controller to shape the compliance of the closed loop SEA.

The chosen compliance shape then determines $C_5(s)$,
\begin{equation}
\frac{\theta_j}{\tau_j}=\frac{s^2 + \tilde B_2 s + \tilde K_2}{J_j s^2(s^2 + \tilde B_2 s + \tilde K_2)+ K_s (s^2 + \tilde B_1 s + \tilde K_1)},\label{eq:total_compliance}
\end{equation}
which is often a primary consideration when choosing the two poles and two zeros of $C_4(s)$.

\subsection{Virtual Motor System}
As shown in Fig.~\ref{fig:schematics}, we treat $S_5$ as a virtual motor for the meta-SEA. However, $S_5$ is fourth order rather than second order, and has a compliance with respect to motor torque, $H_5(s)$, which is different from its compliance with respect to external torques, $C_5(s)$. Therefore, some manipulation is needed to force it to behave like a motor system.

First, we introduce the system $\hat S(s)$ which we intend to make $S_5(s)$ resemble. This desired virtual motor system has the same compliance with respect to external torques and motor torques: $1/(\hat J_m s^2 + \hat B_m s)$.

For our purpose it is sufficient for $C_5(s)$ to only approximately match $\hat C_5(s)$ at low frequencies. We apply the following approximation of \eqref{eq:total_compliance},
\begin{equation}
\hat C_5=\frac{1}{K_s/\tilde K_2 \cdot (s^2 + \tilde B_1 s + \tilde K_1)},
\end{equation}
which is a reasonable estimate for angular frequencies satisfying $\tilde K_2 \gg \|s^2+\tilde B_2 s\|$ and for $\hat J_m$ such that $\hat J_m\gg J_e$. This relationship then determines parameters of the SEA compliance $C_4$: $\tilde K_2 =K_s/\hat J_m $, $\tilde K_1=0$, and $\tilde B_1 = \tilde K_2/K_s \cdot \hat B_m$. 

Compared to accurately matching $C_5(s)$ to $\hat C_5(s)$, compensating for the behavior of $H_5(s)$ is very important. This is accomplished with a special purpose filter $G_v(s)$,
\begin{equation}
G_v(s) = J_m/\hat J_m \left[1 + \frac{J_j}{K_s}\frac{s^2(s^2 +\tilde B_2 s + \tilde K_2)}{s^2 + \tilde B_1 s + \tilde K_1}\right],
\end{equation}
which we use to define a new virtual motor torque variable $\hat \tau_m$, s.t. $G_v \hat \tau_m = \tau_3$. And with this compensator, we can reasonably expect the transfer function $\theta_j/\hat{\tau_m}$ to be $1/(\hat J_m s^2 \hat B_m s)$. However, in order to implement this non-causal $G_v(s)$ filter, we will need to introduce low pass filters. These, in conjunction with the feedback delay, contribute phase lag to this transfer function in practice.

\subsection{Compliance Ratios and the Amplification Rate}
In our linear model, the joint angle of an exoskeleton which is in contact with both an amplified operator and an environment is the superposition of the two joint signals from the operator and environment. In the special case where the output motion is locked,
\begin{gather}
\theta_j = 0 = C_5 \tau_{\mathrm{environmnet}}+C_6 \tau_{\mathrm{human}},\\
\tau_{\mathrm{environment}} = -C_6/C_5 \tau_{\mathrm{human}}.
\end{gather}
That is to say, the ratio $C_6(s)/C_5(s)$ is the amplification of the human. If both $C_5$ and $C_6$ are minimum phase and stable, then this amplification will also be minimum phase and stable. Note that the cuff spring is irrelevant to this ratio.

\section{Transmission Disturbance Observer}

The motor dynamics have been, up to this point in the paper, modeled as a linear system. However, in truth there are highly nonlinear friction effects in the transmission which would prevent easy back-driving of the device under normal circumstances. Let us consider a model of the motor dynamics alone which treats this deviation from linearity as a torque disturbance $\delta_f$:
\begin{equation}
\tau_s+\tau_m +\delta_f = J_m \ddot \theta_m + B_m \dot \theta_m.\label{eq:dob_eq}
\end{equation}
We call this the autonomous motor model, since it has no relationship to the unknown environment. Allowing the controller to treat the transmission as an autonomous system is a lesser known advantage of SEAs.

A recent advancement in SEA control is the disturbance observer (DOB), which estimates and then compensates for some unsensed (disturbance) input to the plant. Different cascade control structures result in widely varying DOB behaviors, as these cascade structures have different SISO plants in their design models. DOB structures applied to SISO models of motor position \cite{KongBaeTomizuka2009TMech} cancel the combination of friction and back-driving forces that disturb the otherwise current-driven motor model. When applied to rejecting disturbances in a closed-loop force-control plant, the DOB must cancel out the transmission friction and output acceleration inputs which act alongside the force set-point to influence the spring force \cite{PaineOhSentis2014TMech}. It is also possible to directly reject disturbances in the open loop model \cite{HopkinsEA2015IROS,LuHaningerChenTomizuka2015AIM}, which has shown to be an equivalent approach \cite{RoozingWalzahnCaldwellTsagarakis2016IROS}. These force disturbance observers improve impedance rendering accuracy when used in a cascaded impedance controller \cite{MehlingHolleyOMalley2015IROS}. 

As they relate to understanding the limits of SEA performance, DOBs have two distinct effects. When a DOB cancels transmission friction, it works against the largest source of nonlinearity in the system and therefore makes linear stability analysis more accurate. On the other hand, when the DOB attempts to cancel an important aspect of the system behavior it can make the system more difficult to stabilize. For instance, canceling the relationship between output acceleration and spring force is an implicit acceleration feedback, which can jeopardize the passivity of the output (and require a minimum output inertia for stability). We therefore propose a transmission DOB which only removes errors in the autonomous motor model. This should make our system behave more linearly, while leaving any drastic changes to the system behavior to the linear feedback controller designed via compliance shaping.

\begin{figure}
	\centering\resizebox{\columnwidth}{!}{{\footnotesize%
		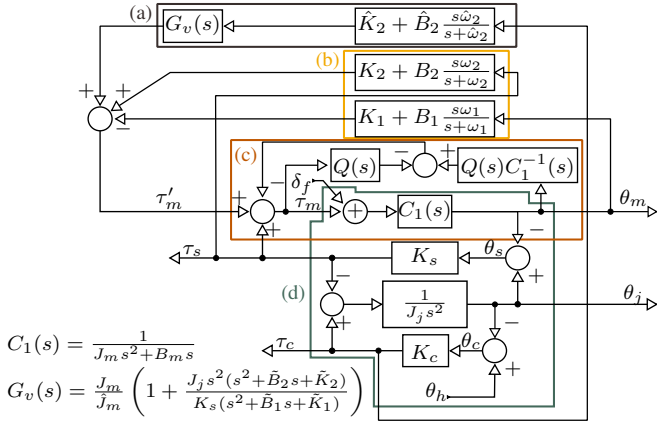}}
	\caption{Block Diagram showing a) the cuff torque compensator, b) the SEA compliance shaping controller, c) the disturbance observer (along with the system it acts on), and d) our model of the physical system. .}\label{fig:BlockDiagram}
\end{figure}
The block diagram in Fig.~\ref{fig:BlockDiagram} shows our whole controller and the physical system (d). At the top level, (a) shows the force feedback control for the virtual SEA, with the filter $G_v(s)$ that compensates for the difference between motor torques and forces on the exoskeleton. A SEA compliance shaping controller (b) uses motor and spring torque feedback to make $C_5(s)$ resemble $\hat C_5(s)$. The disturbance observer (DOB) is the lowest level of control and only seeks to make the linear assumptions more accurate by rejecting the transmission force disturbance $\delta_f$.

\begin{figure}
	\centering{\footnotesize\def\svgwidth{3in}
	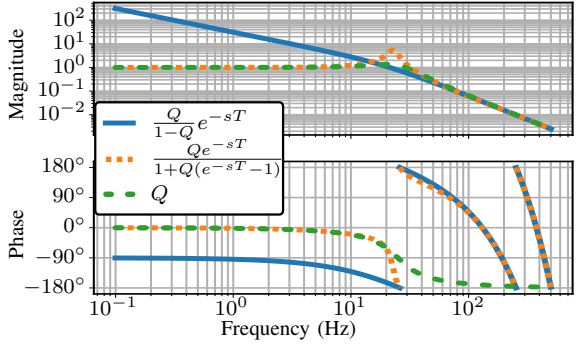}
	\caption{Tuning the disturbance observer's Q filter to avoid destabilization by time delay.}\label{fig:DOBtune}
\end{figure}
The Bode plot in Fig.~\ref{fig:DOBtune} shows three transfer functions which explain the tuning limit of the DOB Q filter. First, looking closely at Fig.~\ref{fig:BlockDiagram}.c, we can see that the DOB has a net feedback gain of $Q(s)/(1-Q(s))C_1^{-1}(s)$ between output $\theta_m$ and input $\tau_m$. If we assume the model of the inverse plant is accurate, then this is essentially an open loop gain of $Q(s)/(1-Q(s))$ from $\tau_m$ back to $\tau_m$. The first line in Fig.~\ref{fig:DOBtune} is this open loop gain multiplied by the system's control time delay (essentially, an un-modeled aspect of the plant). When this open loop is closed, we can see how the time delay distorts the poles of the low pass filter Q---the second line in Fig.~\ref{fig:DOBtune}. The third line is the original Q filter, for comparison. When a Q filter is chosen with an excessive bandwidth or insufficient damping, this second line will reveal unstable pole behavior in the Bode plot.

\section{Double Compliance Shaping Control Design}\label{sec:theory}

\begin{figure*}
	\centering{{\footnotesize\def\svgwidth{6.0in}
		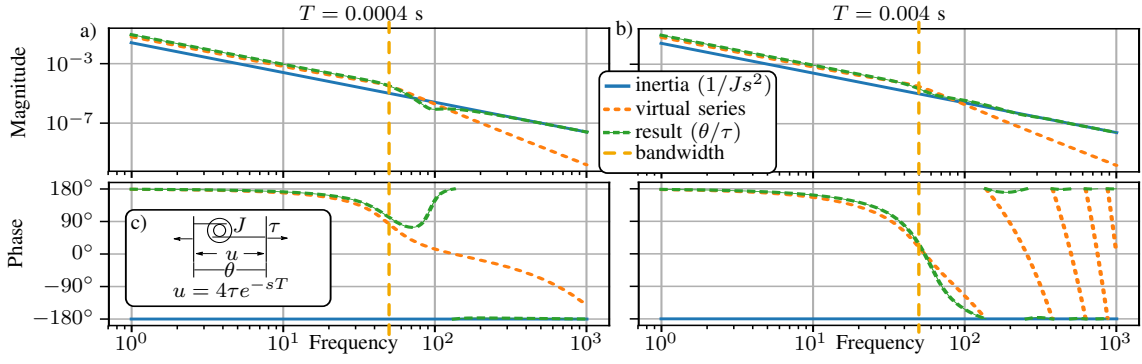}}\caption{For the simple system in (c), amplification control is shown for two different time delay values. The bandwidth limitation (which inevitably exists) is modeled as a second order low pass filter at 50 Hz. In (a) the behavior is minimum phase and stable. In (b), the larger time delay  leads to a non-minimum phase zero and an unstable impedance. The fundamental difficulty of amplification is that even if instability is avoided, there is no way to avoid a phase which drops below the passive phase region.}\label{fig:amplification}
\end{figure*}

Fig.~\ref{fig:amplification} demonstrates a fundamental stability issue of strength amplification exoskeletons: in order to increase compliance, a sort of ``phase-debt'' must be paid eventually. Unlike the introduction of non-minimum phase zeros, this problem cannot be avoided. When a human operator acts as a spring of sufficient stiffness, this non-passive phase can lead to instability. For this reason, we've added a mechanical spring in series with the cuff. While this will not resolve the issue of passivity, it helps ensure coupled stability by switching the problem of destabilization by stiff springs to the problem of destabilization by light inertias.

\begin{figure}
	\centering{\footnotesize\resizebox{\columnwidth}{!}{%
	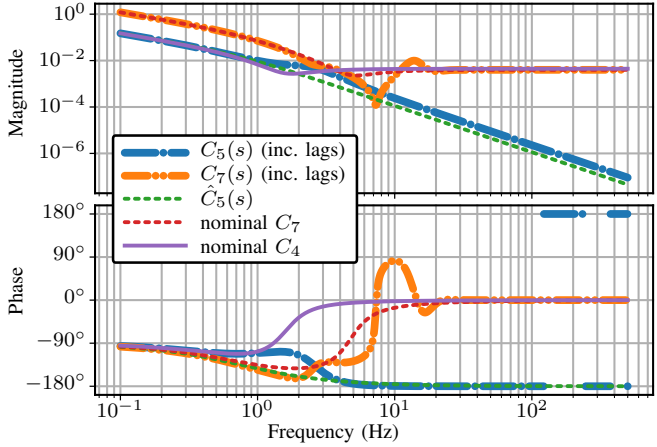}}
	\caption{Bode plot of the amplification controller design.}\label{fig:design}
\end{figure}

\begin{figure}
	\centering{\footnotesize\resizebox{\columnwidth}{!}{%
		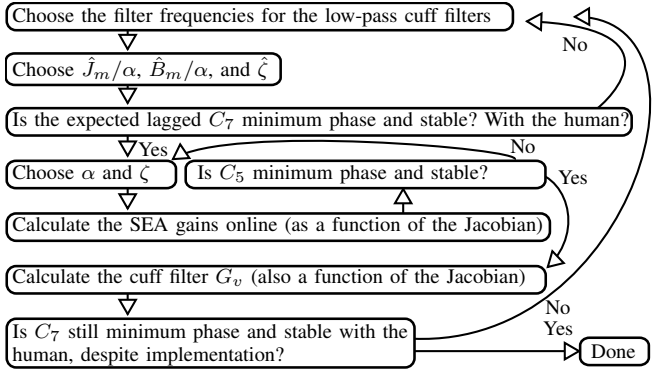}}
	\caption{Flow chart of the controller design process.}\label{fig:flowchart}
\end{figure}

The process we use to design the controller is visualized in Fig.~\ref{fig:design} and Fig.~\ref{fig:flowchart}. Our approach first designs a nominal compliance for $C_7$ (the compliance of the exoskeleton's cuff angle with respect to human torques) and then for $C_5$ (the compliance of the exoskeleton angle with respect to external torques from the environment). These nominal compliances are only approximations. In the case of $C_7$ this is due to the many low pass filters which must be added to the cuff torque signal in order to actually implement the compensator $G_v$ (which is not causal), as well as the time delay. For $C_5$, it is because the nominal motor model is only second order, while the actual system is fourth order and does not have any ability to alter its high-frequency inertia behavior. However, these nominal transfer functions are still very useful since they allow us to parameterize the controller in terms of nominal poles and zeros, asymptotic behaviors, or damping ratios for pole and zero pairs.

The first part of our design process (the top of Fig.~\ref{fig:flowchart}) is to set the filter frequencies and damping ratios for the low pass filters required for implementing the cuff torque compensation filter $G_v$. This is a straightforward trade-off between noise amplification and phase lag. The filters will be used to sanity-check the shape of $C_7$, as shown in Fig.~\ref{fig:flowchart}. However, the discrete-time implementation of these filters may slightly differ from the behavior of the continuous time filters, and this will require iterative design. Next, the behavior of the nominal cuff compliance $C_7$ is specified using the attenuated virtual motor inertia $\hat J_m/\alpha$, damping $\hat B_m/\alpha$, and the damping ratio $\hat \zeta$ of the nominal second order zero occurring at the intersection with the cuff spring compliance line. These parameters are chosen before the amplification rate and only describe the nominal $C_7$ behavior. 

At this point, we should be able to test the expected $C_7$ (including the low pass filters). Unfortunately, our implementation of the $G_v(s)$ only filters the part of the expression which needs a second order derivative, and this results in an actual filtering behavior which is slightly different (with a slightly better phase) than our prediction step, which assumes that the whole cuff feedback signal is filtered by these low-pass filters.

After choosing the amplification ratio $\alpha$, we can describe the nominal $C_7(s)$,
\begin{equation}
C_7(s)=\theta_h/\tau_c \approxeq \frac{s^2 + 2 \hat \zeta \sqrt {\alpha K_c/\hat J_m}+\alpha K_c/\hat J_m }{K_c (s^2 + \hat B_m/\hat J_m s)},\label{eq:nominal_C7}
\end{equation}
which entails gains 
\begin{gather}
\hat K_2 = \alpha-1,\quad \hat B_2 = (2 \hat \zeta \sqrt{K_c \hat J_m \alpha } - \hat B_m)/K_c,
\end{gather}
following the same logic as Sec.~\ref{sec:compliance_shaping}. At this step, we have designed the controller for the meta-SEA based on its compliance shape \eqref{eq:nominal_C7}. The above gains represent the spring force feedback controller for the meta-SEA. All that remains is to design the SEA compliance $C_4(s)$ such that the joint compliance $C_5(s)$ resembles the desired virtual motor compliance $\hat C_5(s)$ at low frequencies.

There is only one free parameter left---the damping ratio of the second order zero pair $\zeta$. The $C_4(s)$ which yields the appropriate low frequency behavior is
\begin{equation}
C_4(s) = \theta_j/\tau_s = \frac{s^2+2\zeta \sqrt{K_s/\hat J_m}s + K_s/\hat J_m}{K_s(s^2 + \hat B_m/\hat J_m s)}.
\end{equation}
And again, compliance shaping converts this desired compliance into actionable gains
\begin{gather}
K_1 = 0, \quad B_1 = \hat B_m J_m/\hat J_m - B_m,\\
K_2 =  J_m/\hat J_m - 1,\\ B_2 = (2\zeta\sqrt{K_s/\hat J_m} -\hat B_m /\hat J_m)J_m/K_s.
\end{gather}
From here, the filter $G_v$ can be calculated, and there are no more free parameters anywhere in the control.

What this control law is doing, in essence, is first setting the $C_7(s)$ compliance as high as possible without introducing a non-minimum phase zero. Then, in order to accomplish the desired amplification $\alpha$ it reduces the compliance of $C_5(s)$ by using negative spring torque feedback to amplify inertia of the motor, which dominates $C_5(s)$ at low frequencies. This has the unintended side effect of amplifying transmission friction in the output. However, this effect is mitigated by the transmission disturbance observer.

%
%
%This strategy of tuning SEA behavior to achieve a higher inertia by using negative spring feedback to amplify the reflected inertia of the motor is one of the contributions of this paper. It works much better if the spring stiffness is higher.

As the joint angle changes, so does the transmission Jacobian, and with it the reflected motor inertia, motor damping, and spring stiffness. Fortunately, our algebraic method is extremely simple computationally, and we can re-generate the appropriate gains according to the new values of these parameters at every time step. This simple gain scheduling approach comes with no guarantee of stability, but, in practice, is effective. Even as the Jacobian changes, the compliance $C_7(s)$ that the human feels should stay the same. However, the compliance that the environment feels, $C_5(s)$, will change noticeably. When the reflected spring compliance, $1/K_s$, increases, the frequency at which $C_5(s)$ deviates from its nominal model drops. When this happens, the bandwidth of the amplification ratio also decreases.

\subsection{Passivity of the Operator Compliance}
While the nominal operator compliance can be made passive, the phase lag in the cuff feedback signal make it impractical to actually achieve this. As pointed out in \cite{HeHuangThomasSentis2019ArXiv}, the human provides a very convenient hysteretic damping, and therefore using the passivity criteria to judge coupled stability with a human operator is very conservative. We instead claim that the addition of a spring, along with the SEA compliance shaping technique which ensures that the zeros introduced by this spring are sufficiently damped, serves to avoid oscillations due to high-stiffness operators interacting with high-frequency amplification. The strategy simply substitutes a susceptibility to spring-like humans for one to inertia-like humans.

%\begin{figure*}\centering%
%	\input{ResultsPlot.pdf_tex}%
%	\caption{Demonstration of higher than passive stiffness control achieved with compliance shaping. Bode plots show compliance transfer functions for the passive spring, the closed loop actuator, and the closed loop joint---which is inferred from the actuator behavior and the ideal joint inertia model. The actuator is less compliant than the passive spring---and therefore is not passive at high frequencies. Joint compliance at high frequencies is dominated by the inertia, and therefore satisfies a relaxed passivity condition based on a slightly wider phase range. Subplot a) shows the raw frequency domain measurements from 20 experiments using the same controller. Subplot b) shows the same data with a compensatory term that eliminates friction and other errors in the motor model (i.e. corrected motor position is found as a transfer function of the difference between motor torque and spring torque).}\label{fig:results}
%\end{figure*}

\section{Experiments}\label{sec:experiments}

Using the hardware setup pictured in Fig.~\ref{fig:hardware}, we were able to test several iterations of this controller. We were not able to exhaustively explore the parameter space of the controllers, but the stability predictions of the model (when accounting for the time delay, filtering lags, and discrete time implementation of $G_v$) have been reliable thus far.

\begin{figure}\vspace{.1in}\centering\resizebox{\columnwidth}{!}{%
	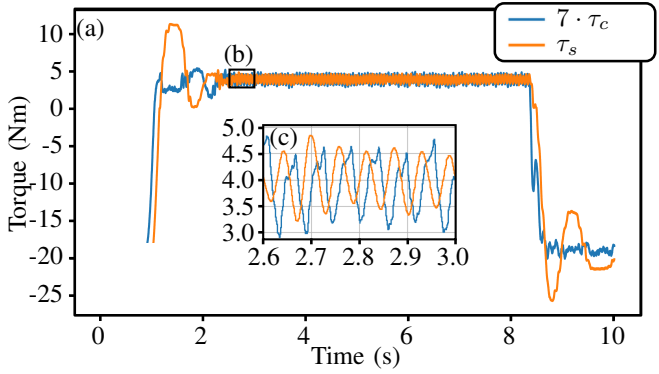}
	\caption{Demonstration of strength amplification behavior with output locked. }\label{fig:timedomain}
\end{figure}

Fig.~\ref{fig:timedomain} shows an experiment where the exoskeleton output is locked, and compares the torque applied by the actuator to the low frequency amplification gain $\alpha-1$. This test illustrates the success of our amplification, since the $\tau_s$ is clearly tracking $7 \cdot \tau_c$. It also confirms the basic premise of our control design process: that high amplification ratios can be obtained by simultaneously increasing the compliance the human feels, $C_7(s)$, and decreasing the compliance the environment feels, $C_5(s)$. While it may be more common to think only of increasing $C_7(s)$ and making the human's life easier, $C_7(s)$ is harder to raise than $C_5(s)$ is to reduce. 

Some residual oscillation (at roughly 17.5 Hz) occurs in steady state, which can be seen in Fig.~\ref{fig:timedomain}.b and in its magnification axes Fig.~\ref{fig:timedomain}.c. This may be due to a slight backlash in the transmission. Negative torques push against the joint limit in this experiment. Imperfect calibration of the 6-axis cuff force/torque sensor zero point leaves the cuff torque signal $\tau_c$ with a roughly 4/7 Nm bias. Together, these points suggest that this vibration occurs near zero force, and may therefore be related to backlash in the transmission.

\begin{figure}\vspace{.1in}
	\centering{\resizebox{\columnwidth}{!}{\footnotesize
		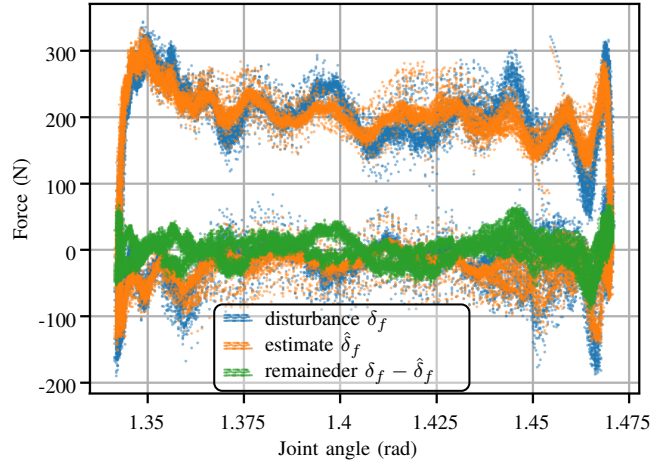}}
	\caption{A test of the disturbance observer, using a simple position controller and a slow sinusoidal position input. This test shows the hysteretic behavior of the transmission disturbance. The disturbance observer estimates the equation error in motor system torque balance, and uses this estimate to cancel the undesirable coulomb friction behavior.}\label{fig:DOBTest}
\end{figure}

To demonstrate the benefit of using the disturbance observer, we conducted a simple experiment with position control SEA gains, no cuff feedback, and a slow sinusoidal input. The actuator forces for this experiment are plotted in Fig.~\ref{fig:DOBTest} against the joint angle. Since this is a repetitive test, we can see how the transmission force is essentially a hysteretic friction effect. The disturbance signal is found off-line by differentiating motor velocity, and computing the equation error in \eqref{eq:dob_eq}. On top of this, we plot the disturbance observer's internal state representing its estimate of this disturbance, and also the difference between the estimate and the equation error. Only this small discrepancy disrupts the linearity of the motor system as it appears to the higher-level of controllers.

%The controller design in Fig.~\ref{fig:design} is stable both in mid-air and when constrained, but can be made to vibrate by placing small objects on the cuff. It does not, however, vibrate when the human stiffens their elbow.

%\begin{table}
%\caption{Parameters of the Actuator and Controller}\centering
%\begin{tabular}{cc cc}
%\toprule 
%Parameter & Value \\
%\midrule
%%$J_m$ & 0.44 Kg m$^2$ \rule{0em}{1.1em} \\
%%$B_m$ & 17.9 Nm s/rad \rule{0em}{1.1em} \\
%%$K_s$ & 1$\,$180 Nm/rad \rule{0em}{1.1em} \\
%%$J_j$ & 0.29 Kg m$^2$ \rule{0em}{1.1em} \\
%\bottomrule
%\end{tabular}
%\begin{tabular}{cc}
%	\toprule
%	Gain & Value \\
%	\midrule
%%	$K_1$ & 69$\,$480 Nm/rad \rule{0em}{1.1em} \\
%%	$B_1$ & 1$\,$387 Nm s/rad \rule{0em}{1.1em} \\
%%	$K_2$ & -41$\,$690 Nm/rad \rule{0em}{1.1em} \\
%%	$B_2$ & -1$\,$183 Nm s/rad \rule{0em}{1.1em} \\
%	\bottomrule
%\end{tabular}
%\end{table}

\section{Conclusion}\label{sec:conclusion}
This paper shows that strength amplification exoskeletons can accomplish their objective equally effectively by increasing the compliance they show to the human or decreasing the compliance they show to the world. We  developed an amplification controller which allows for elastic cuffs by treating the exoskeleton as a series elastic actuator system (the motor of which is also a series elastic actuator). Furthermore, we designed a SEA-specific disturbance observer which, unlike other disturbance observers, does not attempt to remove an input from the nominal plant which could be otherwise handled by compliance shaping.

\bibliographystyle{IEEEtran}
\bibliography{bib}

\end{document}